\documentclass[lettersize,journal]{IEEEtran}
\usepackage{amsmath,amsfonts}
\usepackage{algorithmic}
\usepackage{algorithm}
\usepackage{array}
\usepackage[caption=false,font=normalsize,labelfont=sf,textfont=sf]{subfig}
\usepackage{textcomp}
\usepackage{stfloats}
\usepackage{url}
\usepackage{verbatim}
\usepackage{graphicx}
\usepackage{cite}
\usepackage{multirow}
\begin{document}

\title{DRM-IR: Task-Adaptive Deep Unfolding Network for All-In-One Image Restoration}

	\author{Yuanshuo Cheng, Mingwen Shao, ~\IEEEmembership{Member,~IEEE}, Yecong Wan, and Chao Wang
		\thanks{Yuanshuo Cheng, Mingwen Shao and Yecong Wan are with the College of Computer Science and Technology, China University of Petroleum, China.}
    \thanks{Chao Wang is with the Centre for Artificial Intelligence, University of Technology Sydney, Ultimo, NSW 2007, Australia.}
  }

\markboth{Journal of \LaTeX\ Class Files,~Vol.~14, No.~8, August~2021}%
{Shell \MakeLowercase{\textit{et al.}}: A Sample Article Using IEEEtran.cls for IEEE Journals}


\maketitle

\begin{abstract}
Existing All-In-One image restoration (IR) methods usually lack flexible modeling on various types of degradation, thus impeding the restoration performance. To achieve All-In-One IR with higher task dexterity, this work proposes an efficient Dynamic Reference Modeling paradigm (DRM-IR), which consists of task-adaptive degradation modeling and model-based image restoring. Specifically, these two subtasks are formalized as a pair of entangled reference-based maximum a posteriori (MAP) inferences, which are optimized synchronously in an unfolding-based manner. With the two cascaded subtasks, DRM-IR first dynamically models the task-specific degradation based on a reference image pair and further restores the image with the collected degradation statistics.
Besides, to bridge the semantic gap between the reference and target degraded images, we further devise a Degradation Prior Transmitter (DPT) that restrains the instance-specific feature differences.
DRM-IR explicitly provides superior flexibility for All-in-One IR while being interpretable. Extensive experiments on multiple benchmark datasets show that our DRM-IR achieves state-of-the-art in All-In-One IR.
\end{abstract}

\begin{IEEEkeywords}
All-in-one image restoration, Model-guide design, deep convolutional neural network, Transformer, Task adaptive
\end{IEEEkeywords}

\section{Introduction}
\IEEEPARstart{I}{mage} restoration (IR) aims to recover clean and high-quality images from degraded images with adverse degradations, which is critical for subsequent high-level vision tasks such as object detection \cite{liu2016ssd,redmon2016you,carion2020end} and image segmentation \cite{long2015fully,chen2014semantic,chen2018encoder}. Existing methods based on Convolutional Neural Networks (CNNs) and Transformers have achieved excellent performance on task-specific IR in recent years \cite{mehri2021mprnet,wu2021contrastive,wan2022image,shao2021uncertainty}. But these methods usually train models individually for each specific task, which lowers their practicality and convenience for implementation. 
\begin{figure}[t]
	\begin{center}
		\includegraphics[width=0.98\linewidth]{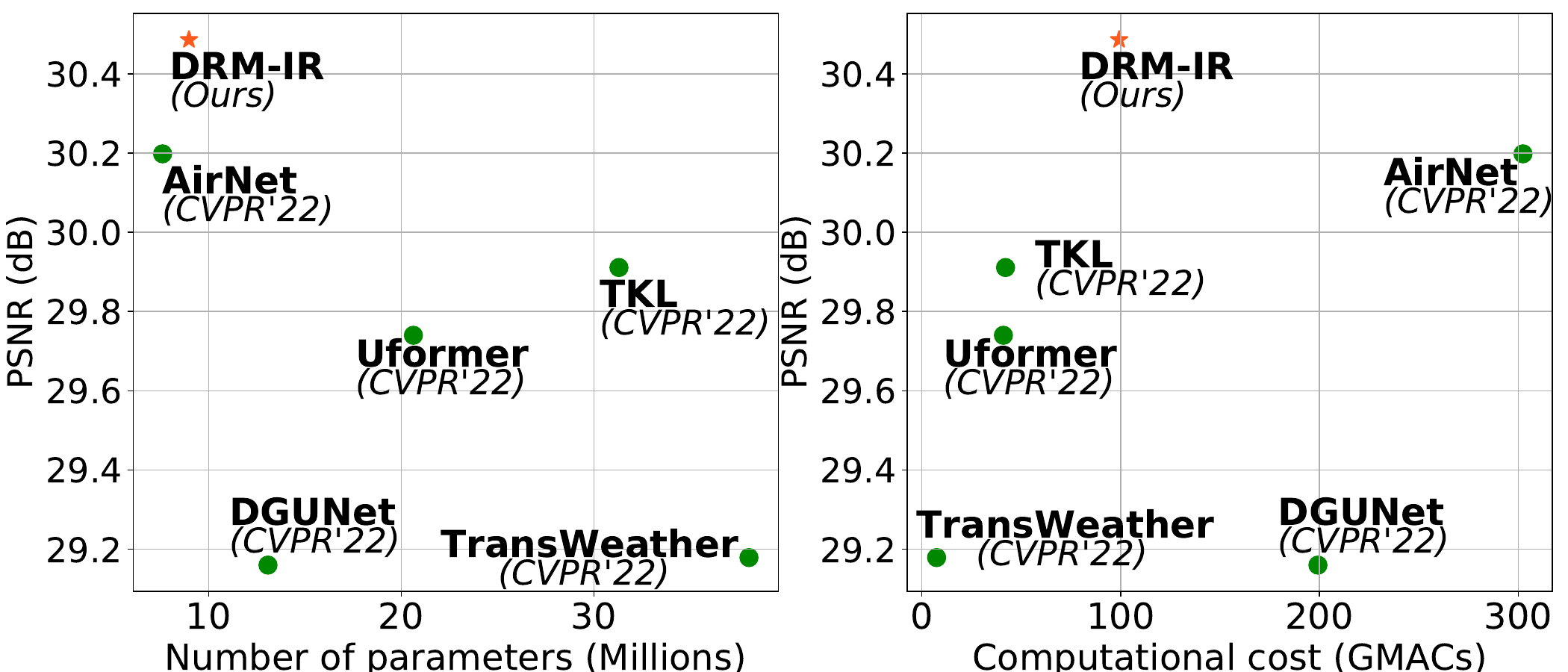}
	\end{center}
 \vspace{-10pt}
	\caption{Average PSNR vs. parameter quantity and computational cost on All-In-One IR. The proposed DRM-IR reaches state-of-the-art performance with competitive computational efficiency.}
	\label{imgscore}
\end{figure}
Aiming at this problem, several learning-based methods have explored All-In-One image restoration. For instance, Li \emph{et al.} first propose to apply a multi-encode structure to deal with multiple degradations \cite{li2020all}. Furthermore, a few subsequent works have attempted to design more concise and powerful models. For example, comparative learning \cite{li2022all}, knowledge distillation \cite{chen2022learning}, and learnable Queries \cite{valanarasu2022transweather} are employed to enhance the adaptability of the end-to-end models for multiple degradations. Nevertheless, these methods tend to directly and impartially learn a general mapping for various degradations within a black box, thus further limiting the All-In-One restoration performance and interpretability.

In contrast, a few interpretable model-based methods \cite{chen2016trainable,kokkinos2018deep,kruse2017learning,sun2016deep} have also been proposed to construct individual degenerative processes respectively for each IR task. Recently, several works \cite{zhang2017learning, zhang2020deep, wu2022uretinex, ding2018domain} have further integrated this model-based idea with CNNs (\emph{e.g.}, Deep Unfolding) to provide a trainable task-specific solution in an end-to-end manner. However, these methods still require one-by-one pre-designed models for each type of degradation, which precludes the All-In-One IR implementation. Moreover, modeling accuracy is critical for model-based methods, but some complex degradations models (\emph{e.g.}, rain streak, haze, low light, etc.) are intractable to construct precisely. To sum up, there is an urgent need for a trade-off solution that combines the convenience of All-In-One restoration with the flexibility of task-specific modeling.

To address the aforementioned challenge, this work innovatively explores the unfolding-based paradigm for All-in-One restoration and proposes an efficient Dynamic Reference Modeling IR method (DRM-IR) with two entangled subtasks: (1) task-adaptive degradation modeling and (2) model-based image restoring. 
The former aims at dynamically modeling specific degradations in the multi-corruption scenario. 
The latter further eliminates the degradation based on the constructed model.

To be specific, for subtask (1), a MAP inference is constructed by introducing a reference image pair to generate the degradation matrices. For subtask (2), another MAP inference is constructed relying on the generated degradation matrices. Then the degradation is eliminated via iterative optimization. Besides, to entangle the two subtasks, we propose a Degradation Prior Transmit (DPT) to bridge the semantic gap between the target to-be-restored image and the external reference image. DPT facilitates the restoration process by explicitly fine-tuning the degradation matrices in coordination with the target image. Meanwhile, a joint optimization framework based on deep unfolding methods is developed to solve these two optimization tasks in parallel.

To the best of our knowledge, this is the first work aiming at bridging unfolding-based methods to the All-In-One IR community. The proposed DRM-IR explicitly enables flexible treatment for different types of degradations via task-adaptive modeling, which promotes the unified elimination of multiple degradations while being competitive in terms of the number of parameters and computational cost (see Fig. \ref{imgscore}). Notably, this work transforms the uninterpretable adaptation problem for different degradations in existing All-In-On IR into a specific problem of modeling different degradation matrices, which achieves interpretable flexibility in multi-degradation scenarios. In addition, the MAP inference-based solution framework also enables the model to inherit favorable interpretability. In summary, the main contributions of this work can be summarized as follows:
\begin{itemize}\setlength{\itemsep}{0.5pt}
	\item A novel framework for All-In-One image restoration called DRM-IR is developed, which intuitively treats this inverse problem as two coupled subtasks: task-adaptive modeling and model-based image restoring. This paradigm ensures superior flexibility in multi-degradation scenarios while being interpretable.
	\item A reference-based task-adaptive degradation modeling method is proposed, which implements an adaptive construction for different degradation matrices while improving the modeling accuracy by introducing additional external reference image pairs.
	\item A degradation prior transmitter is devised to further bridge the semantic gap between the target and reference images, which consequently entangles the proposed two subtasks in a unified framework.
	\item Extensive experiments on multiple benchmark datasets demonstrate that the proposed method achieves state-of-the-art performance on All-In-One and Task-specific IR.
\end{itemize}

\begin{figure*}
\begin{center}
	\includegraphics[width=\linewidth]{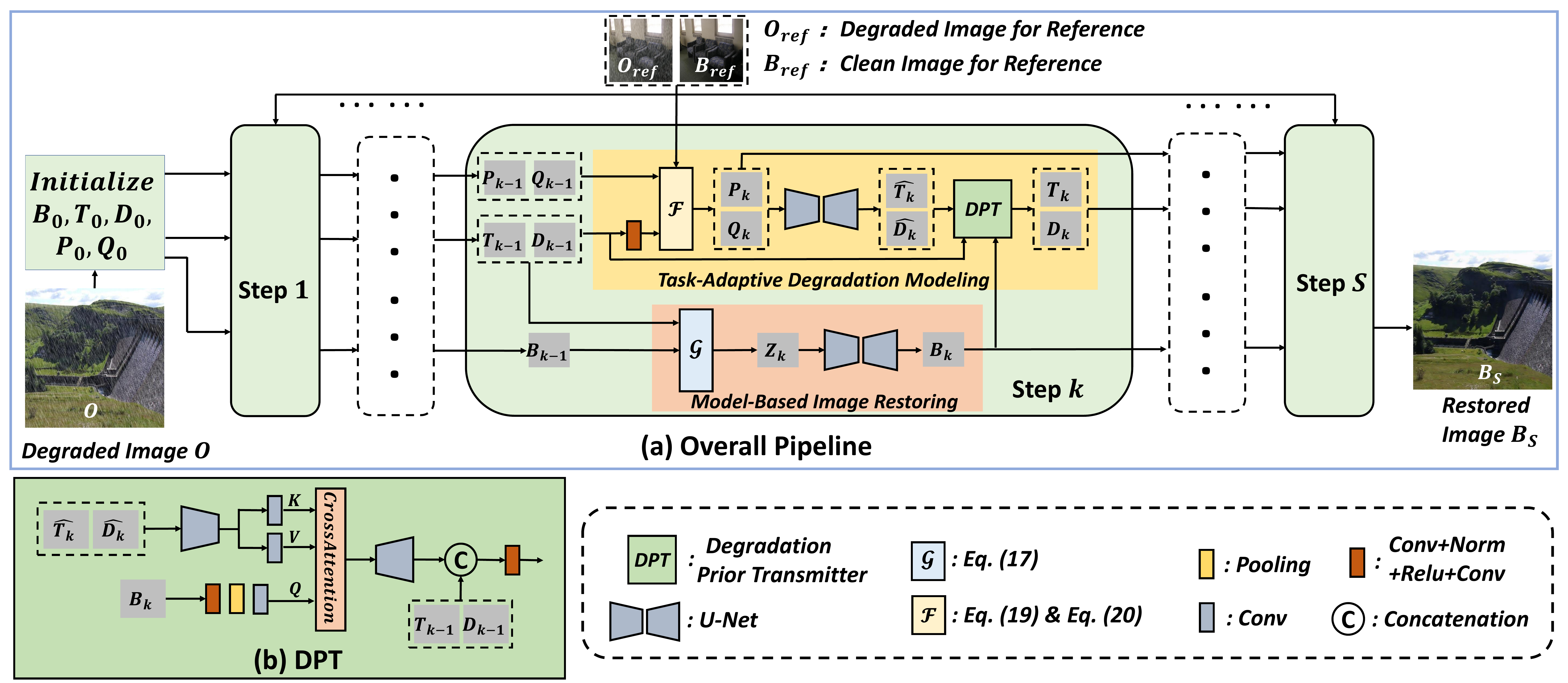}
\end{center}
\caption{Illustration of the proposed DRM-IR. (a) The overall pipeline of DRM-IR. Where T and D denote the degradation matrices and P and Q are their corresponding auxiliary variables. (b) Details of DPT. The DPT is devised to introduce additional degradation priors from the reference image pair into the degradation modeling process. Please zoom in to see the details.}
\label{imgframework}
\end{figure*}

\section{Related work}
\label{2}
\subsection{All-In-One Image Restoration}
\label{2.1}
To overcome the cumbersomeness caused by requiring separate training models for each degradation in task-specific IR, several works in recent years have proposed to design a unified model to dismantle multiple degradations. The key to All-In-One image restoration lies in enabling the model to adapt flexibly to different types of degradations. For instance, Zhu et al. \cite{zhu2023learning} argued that various weather-related degradations encompass both general and specific characteristics. To address this, they proposed a regularization-based optimization strategy to extend weather-specific parameters. Furthermore, some methods employ specialized structures such as dynamic architecture \cite{yang2023visual,zhang2023ingredient} and learnable queries \cite{valanarasu2022transweather} to handle different degradation types. Additionally, certain approaches focus on the training process by incorporating contrastive learning \cite{li2022all} or knowledge distillation \cite{chen2022learning} to enhance the model's ability to remove various types of degradations. However, these frameworks still lack flexibility in the face of complex scenarios with multiple degradations compared to model-based approaches. In contrast, this work non-trivially integrates learning-based and model-based methods to achieve better flexibility and performance than existing literature, as well as better interpretability.

\subsection{Deep Unfolding Method}
\label{2.2}
Zhang \emph{et al.} \cite{zhang2017learning} first pioneer the employment of the Deep Unfolding framework to integrate CNN with model-based methods on task-specific IR tasks. For the degradation model $y=Hx$, the expected clean image can be estimated by minimizing the following energy function:
\begin{equation}
\hat{x} = \arg\mathop{\min}\limits_{x} \frac{1}{2}||y-Hx||^2+\lambda\Phi(x), \label{equnfolding}
\end{equation}
where $y$ is the degraded image, $x$ is the optimized variable, $H$ is the degradation matrix and $\lambda$ is a trade-off parameter. 
$\frac{1}{2}||y-Hx||^2$ is the data term (fidelity term) that constrains the mapping of $x$ to $y$ to conform to the degradative process. $\lambda\Phi(x)$ is the prior term (regularization term) that constrains the solution space by imposing a prior for $x$. 
By applying the HQS algorithm \cite{geman1995nonlinear}, Eq. \ref{equnfolding} can be split into two subproblems focusing on data term and prior term, respectively. 
The optimization of Eq. \ref{equnfolding} can be achieved by iterating these two subproblems alternately. 
The data term subtask is formulated as a simple least squares problem, while the prior term one can be solved via a trainable CNN model.

Following unfolding-based methods\cite{mou2022deep,zhang2017learning,wu2022uretinex,zhang2020deep,jin2022shadowdiffusion} have yielded promising results on task-specific IR tasks such as image denoising, low-light image enhancement (LLIE), shadow removal and etc.
Despite of the flexibility and interpretability on single-task degradation modeling, these methods still cannot achieve a unified All-In-One restoration due to the fact that the degradation model for each corruption type varies and requires pre-designed priors.
Aiming at this limitation, this work further level up the flexibility to All-In-One scenario and introduce a reference-based task-adaptive modeling paradigm DRM-IR. A sophisticated yet efficient All-In-One IR framework is further designed by jointing task-adaptive degradation modeling and model-based image restoring.

\subsection{Reference-based Image Restoration}
\label{2.3}
Reference-based methods have been widely applied in image super-resolution\cite{cao2022reference,huang2022task,pesavento2021attention,xie2020feature}, in which a external high-resolution (HR) reference image is usually utilized to tackle the ill-posed problem as additional information with more fine-grained details.
For example, the classical CrossNet \cite{zheng2018crossnet} aligns HR reference images and low-resolution (LR) images through a optical flow mechanism, which in turn conveys HR texture features. 
In this scheme, we innovatively introduce the reference-based idea into the All-In-One IR task. On the one hand, benefiting from a reference image pair, an interpretable MAP inference is constructed to achieve task-adaptive degradation modeling. On the other hand, additional information is introduced to further enhance the modeling precision, thus leading to higher restoration performance.

\section{Methodology}
\label{3}
The overall pipeline of the proposed DRM-IR can be described as Fig. \ref{imgframework}(a) and Algorithm \ref{alDRM-IR}. In this section, we first formulate the All-In-One IR problem. Afterward, the derivation of the model-based optimization formulation and the initialization for the variables are presented. Thereafter, each key functional module is explained in detail. Finally, the training process and the loss functions are delineated.

\subsection{Problem Formulation}
\label{3.1}
According to previous literature \cite{huang2012context,he2010single,liu2021retinex}, common degradative processes, such as rainy, hazy, and low-light, can be respectively modeled as Eqs. \ref{eqrainy}, \ref{eqhazy}, and \ref{eqlowlight}:

\begin{equation}
O=B+R, \label{eqrainy}
\end{equation}
\begin{equation}
O = TB+(1-T)A, \label{eqhazy}
\end{equation}
\begin{equation}
O = IB, \label{eqlowlight}
\end{equation}
where $O$ and $B$ represent the degraded and clean image. $R$ denotes the rain streak map, $T$ indicates the transmission map, $A$ represents the atmospheric light map and $I$ is the illumination map. Throughout this paper, multiplication operations in formulas are element-wise multiplication unless otherwise specified. With the modeling mentioned above, the universal degradative process can be modeled as:
\begin{equation}
O = TB + D, \label{eqalldeg}
\end{equation}
where $T$ and $D$ denote the transmission map and the degradation map. For convenience, $T$ and $D$ are collectively referred to as the degradation matrices in this paper.

According to the MAP framework, with given degradation matrices (\emph{i.e.}, $T$ and $D$), the IR task can be solved by minimizing the following energy function:
\begin{equation}
\hat{B} =\arg\mathop{\min}\limits_{B}\frac{1}{2}||O-(TB+D)||_F^2+\lambda\Phi(B), \label{eqenergyfres}
\end{equation}
where $||*||_F$ denotes Frobenius norm, $\frac{1}{2}||O-(TB+D)||_F^2$ represents the data term, $\Phi(B)$ is the prior term, and $\lambda $ is a trade-off parameter.

It can be inferred from Eq. \ref{eqenergyfres} that by specifying different degradation matrices, the algorithm can solve a variety of degradation removal problems. Based on this principle, we propose a reference-based approach to adaptively establish different degradation matrices to achieve All-In-One IR, which is the core idea of this work.

Our task-adaptive degradation modeling follows a similar HQS algorithm pattern \cite{geman1995nonlinear} with a non-trivial reference pair. The corresponding degradation matrices can be optimally obtained by minimizing the following energy functions with a given reference image pair:
\begin{equation}
\hat{T},\hat{D} =\arg\mathop{\min}\limits_{T,D}\frac{1}{2}||O_{ref}-(TB_{ref}+D)||_F^2+\mu\Psi(T,D), \label{eqenergyfdeg}
\end{equation}
where $O_{ref}$ and $B_{ref}$ represent the degraded image and clean image from the referenced image pair. $\mu$ donates a trade-off parameter. $\Psi(T, D)$ is the prior term for constraining the solution space of the degenerate matrices.

Based on the deep unfolding framework~\cite{zhang2017learning}, Eq. \ref{eqenergyfres} and Eq. \ref{eqenergyfdeg} are optimized in parallel to achieve reference-based adaptive degradation modeling and established-model-based All-In-One IR. Noticeably, the proposed method holds superior flexibility and interpretability compared to existing All-In-One IR methods.

\subsection{Deep Unfolding-based Optimization}
\label{3.2}
\noindent\textbf{Optimization of model-based image restoring (Eq. \ref{eqenergyfres}).} According to the HQS algorithm \cite{geman1995nonlinear}, the image restoration task can be solved by splitting Eq. \ref{eqenergyfres} into two subproblems and optimizing them alternately. Specifically, by introducing the auxiliary variable $Z$, Eq. \ref{eqenergyfres} can be reformulated as the following optimization problem:
\begin{equation}
\mathop{\min}\limits_{B}\frac{1}{2}||O-(TZ+D)||_F^2+\lambda\Phi(B),\text{ s.t. }Z=B. \label{eqenergyfrescon}
\end{equation}
To handle the equality constraint, we further reformulate it in the following form:
\begin{equation}
\mathop{\min}\limits_{B}\frac{1}{2}||O-(TZ+D)||_F^2+\lambda\Phi(B)+\frac{\gamma}{2}||Z-B||_F^2, \label{eqenergyfresnocon}
\end{equation}
where $\gamma$ is a penalty parameter. The solution to Eq. \ref{eqenergyfresnocon} can be achieved by solving the following two subproblems:
\begin{equation}
Z_k = \arg\mathop{\min}\limits_{Z}\frac{1}{2}||O-(TZ+D)||_F^2+\frac{\gamma}{2}||Z-B_{k-1}||_F^2, \label{eqminz}
\end{equation}
\begin{equation}
B_k = \arg\mathop{\min}\limits_{B}\lambda\Phi(B)+\frac{\gamma}{2}||Z_k-B||_F^2. \label{eqminb}
\end{equation}
Eq. \ref{eqminz} is a least-squares problem, where its solution formula Eq. \ref{eqsolz} can be obtained by differentiating it with respect to $Z$ and setting the derivative to $0$:
\begin{equation}
Z_k = \frac{TO+\gamma B_{k-1}-TD}{\gamma+T^2}. \label{eqsolz}
\end{equation}
For Eq. \ref{eqminb}, which involves a prior term, a learning-based model is employed to learn the clean image prior in a data-driven manner. Specifically, the solution formula of Eq. \ref{eqminb} can be expressed as Eq. \ref{eqsolb}:
\begin{equation}
B_k = M(Z_k,\theta_B), \label{eqsolb}
\end{equation}
where $M$ is a U-Net \cite{ronneberger2015u} providing clean image priors with parameters $\theta_B$. Adopting a CNN to learn prior information from extensive data not only avoids the cumbersomeness of designing priors manually but also enables a more accurate representation of complex information, thus enabling the removal of complex degradations.

\begin{algorithm}[t]
	\caption{Overall Pipeline of DRM-IR}
	\begin{algorithmic}
		\STATE \textbf{Input:} Degraded image $O\in \mathbb{R}^{3 \times H \times W}$, Referenced image pair $O_{ref}\in \mathbb{R}^{3 \times H \times W}$ and $B_{ref}\in \mathbb{R}^{3 \times H \times W}$
		\STATE \textbf{Output:} Clean image $B_s\in \mathbb{R}^{3 \times H \times W}$
		\STATE Initialize $B_0$, $T_0$, $D_0$, $P_0$ and $Q_0$ by Eq. \ref{eqinitb}, \ref{eqinittd} and \ref{eqinitpq}
		\STATE \textbf{for} $k = 1 \to S$\textbf{:}
		\STATE \hspace{0.5cm} update $Z_k$ by Eq. \ref{eqsolzj}
		\STATE \hspace{0.5cm} update $B_k$ by Eq. \ref{eqsolbj}
		\STATE \hspace{0.5cm} \textbf{if} k$<S$\textbf{:}
		\STATE \hspace{1cm} update $P_k$ by Eq. \ref{eqsolpj}
		\STATE \hspace{1cm} update $Q_k$ by Eq. \ref{eqsolqj}
		\STATE \hspace{1cm} update $\hat{T_k}$ and $\hat{D_k}$ by Eq. \ref{eqsoltdj1}
		\STATE \hspace{1cm} update $T_k$ and $D_k$ by Eq. \ref{eqsoltdj2}
		\STATE \hspace{0.5cm} \textbf{end if}
		\STATE \textbf{end for}
		\STATE Return $B_s$
	\end{algorithmic}
	\label{alDRM-IR}
\end{algorithm}

\noindent\textbf{Optimization of task-adaptive degradation modeling (Eq. \ref{eqenergyfdeg}).} The proposed referenced-based degradation modeling also utilizes the HQS algorithm \cite{geman1995nonlinear} with additional reference images. 
The degradation matrices $T$ and $D$ can be obtained by minimizing the Eq. \ref{eqenergyfdeg}. By introducing the auxiliary variables $P$, and $Q$ for $T$, and $D$, Eq. \ref{eqenergyfdeg} can be rewritten as the following optimization problem:
\begin{equation}
	\begin{aligned}
		\mathop{\min}\limits_{T,D}\frac{1}{2}||O_{ref}-(PB_{ref}+Q)||_F^2+\mu\Psi(T,D), \\ \text{s.t. }P=T,Q=D.\label{eqenergyfdegcon}
	\end{aligned}
\end{equation}
To solve the above equation, two quadratic penalty terms are introduced to transform Eq. \ref{eqenergyfdegcon} into the following form:
\begin{equation}
	\begin{aligned}
		\mathop{\min}\limits_{T,D}\frac{1}{2}||O_{ref}-(PB_{ref}+Q)||_F^2+\mu\Psi(T,D)\\+\frac{\alpha}{2}||P-T||_F^2+\frac{\beta}{2}||Q-D||_F^2,\label{eqenergyfdegnocon}
	\end{aligned}
\end{equation}
where $\alpha$ and $\beta$ are penalty parameters. By applying the HQS algorithm \cite{geman1995nonlinear}, Eq. \ref{eqenergyfdegnocon} can be optimized by solving the following three subproblems:
\begin{equation}
	\begin{aligned}
		P_k=\arg\mathop{\min}\limits_{P}\frac{1}{2}||O_{ref}-(PB_{ref}+Q_{k-1})||_F^2\\+\frac{\alpha}{2}||P-T_{k-1}||_F^2,\label{eqminp}
	\end{aligned}
\end{equation}
\begin{equation}
	\begin{aligned}
		Q_k=\arg\mathop{\min}\limits_{Q}\frac{1}{2}||O_{ref}-(P_{k-1}B_{ref}+Q)||_F^2\\+\frac{\alpha}{2}||Q-D_{k-1}||_F^2,\label{eqminq}
	\end{aligned}
\end{equation}
\begin{equation}
	\begin{aligned}
		T_k,D_k=\arg\mathop{\min}\limits_{T,D}\mu\Psi(T,D)+\frac{\alpha}{2}||O_k-T||_F^2\\+\frac{\beta}{2}||Q_k-D||_F^2.\label{eqmintd}
	\end{aligned}
\end{equation}
For the least squares problems 	Eq. \ref{eqminp} and Eq. \ref{eqminq}, the closed-form solutions are calculated directly, and the iterative formulas Eq. \ref{eqsolp} and Eq. \ref{eqsolq} for $P$ and $Q$ are obtained:
\begin{equation}
	\begin{aligned}
		P_k=\frac{O_{ref}B_{ref}+\alpha T_{k-1}-Q_{k-1}B_{ref}}{B_{ref}^2+\alpha},\label{eqsolp}
	\end{aligned}
\end{equation}
\begin{equation}
	\begin{aligned}
		Q_k=\frac{O_{ref}+\beta D_{k-1}-P_{k-1}B_{ref}}{\beta +1},\label{eqsolq}
	\end{aligned}
\end{equation}
The solution of Eq. \ref{eqmintd}, which involves a prior term, can be obtained via a learning-based model as:
\begin{equation}
	\begin{aligned}
		T_k,D_k=M(P_k,Q_k,\theta_{TD}),\label{eqsoltd}
	\end{aligned}
\end{equation}
where $M$ represents a U-Net \cite{ronneberger2015u} and $\theta_{TD}$ donates its parameters.


\noindent\textbf{Joint optimization.} The foregoing two optimization problems are combined to form a coupled system which are optimized jointly to enable flexible All-In-One IR (as illustrated in Fig. \ref{imgframework}(a)). 
On the basis of Eqs. \ref{eqsolz}, \ref{eqsolb}, \ref{eqsolp}, \ref{eqsolq} and \ref{eqsoltd}, our joint optimization formulation can be expressed as follows:
\begin{equation}
Z_k = \frac{T_{k-1}O+\gamma B_{k-1}-T_{k-1}D_{k-1}}{\gamma+T_{k-1}^2}. \label{eqsolzj}
\end{equation}
\begin{equation}
B_k = M(Z_k,\theta_B), \label{eqsolbj}
\end{equation}
\begin{equation}
\begin{aligned}
P_k=\frac{O_{ref}B_{ref}+\alpha T_{k-1}-Q_{k-1}B_{ref}}{B_{ref}^2+\alpha},\label{eqsolpj}
\end{aligned}
\end{equation}
\begin{equation}
\begin{aligned}
Q_k=\frac{O_{ref}+\beta D_{k-1}-P_{k-1}B_{ref}}{\beta +1},\label{eqsolqj}
\end{aligned}
\end{equation}
\begin{equation}
\begin{aligned}
\hat{T_k},\hat{D_k}=M(P_k,Q_k,\theta_{TD}),\label{eqsoltdj1}
\end{aligned}
\end{equation}
\begin{equation}
\begin{aligned}
T_k,D_k=DPT([\hat{T_k},\hat{D_k}],[T_{k-1},D_{k-1}],B_k),\label{eqsoltdj2}
\end{aligned}
\end{equation}
where $[*]$ represents concatenation along the channel dimension. DPT denotes the proposed Degradation Prior Transmitter for bridging the semantic gap between the reference image and the to-be-restored degraded image. The DPT will be explained in detail in Sec. \ref{3.4}.
Besides, in practice, a Convolution-Normalization-Relu-Convolution structure (\emph{i.e.}, CNRC in Fig. \ref{imgframework} (a)) is incorporated to buffer the inter-domain differences between the two tasks.

\noindent\textbf{Variable initializations.}
As described above, for the joint optimization process, the variables that need to be initialized contain the initial clean image $B_0$, the initial degradation matrices $T_0$ and $D_0$, and their corresponding auxiliary variables $P_0$ and $Q_0$.
Since $B_k$ is the optimization objective, we directly initialize it to the degraded image $O$ as:
\begin{equation}
\begin{aligned}
B_0=O.\label{eqinitb}
\end{aligned}
\end{equation}

\begin{figure}
	\begin{center}
		\includegraphics[width=0.95\linewidth]{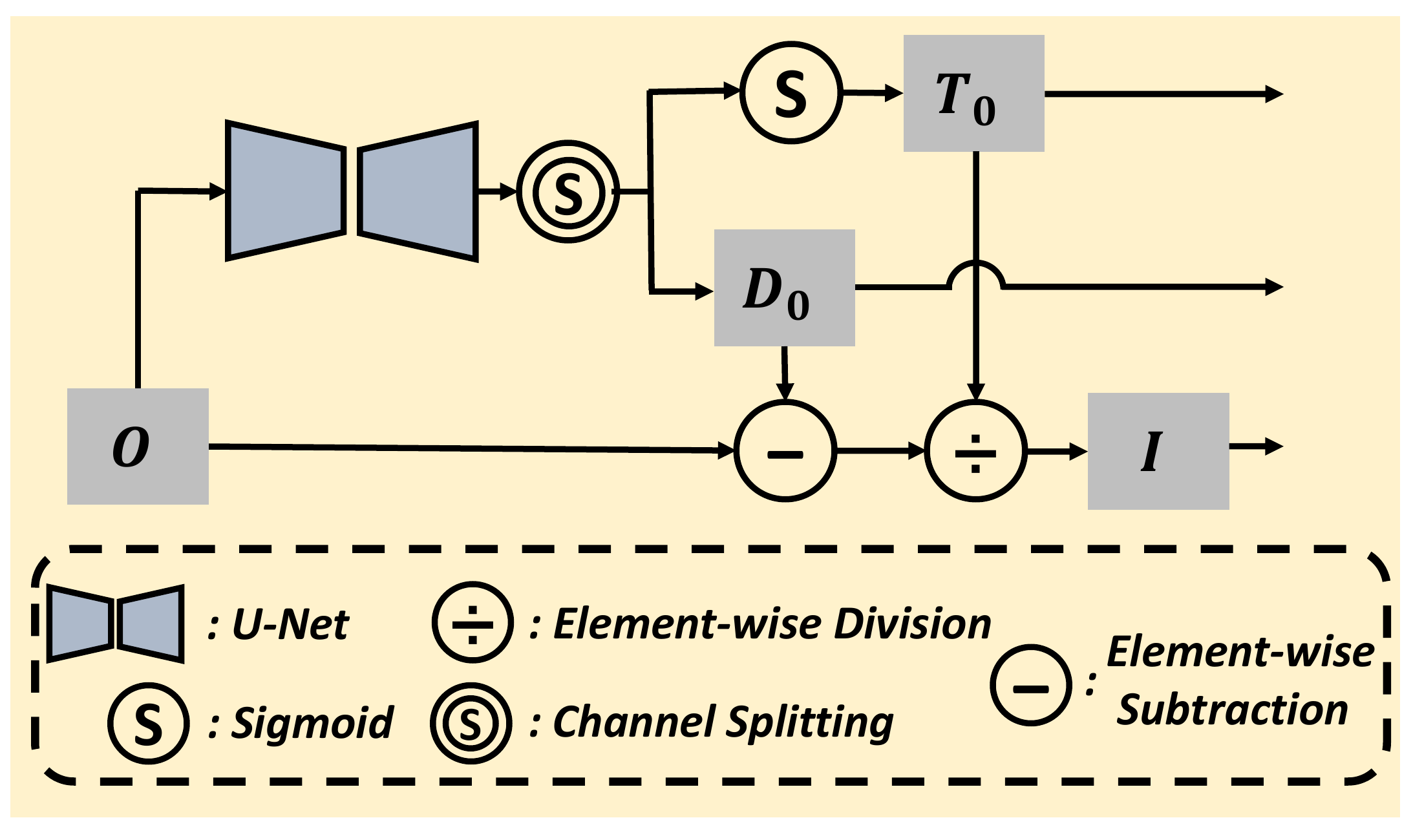}
	\end{center}
  \vspace{-10pt}
	\caption{Details of the estimation for $T_0$ and $D_0$.}
	\label{imgADE}
\end{figure}

Additionally, a pre-trained U-Net \cite{ronneberger2015u} is employed to estimate the cursory initial degradation matrices $T_0$, $D_0$ as:
\begin{equation}
\begin{aligned}
T_0^\prime,D_0&=Split(UNet(O)),\\
T_0 &= Sigmoid(T_0^\prime).
\label{eqinittd}
\end{aligned}
\end{equation}
where $Split$ donates splitting along channel dimension. 
Specifically, as illustrated in Fig. \ref{imgADE}, a cursory clean image $I$ can be obtained by the following equation:
\begin{equation}
\begin{aligned}
I &= (O-D_0)\div (T_0+\epsilon),
\label{eqade}
\end{aligned}
\end{equation}
where $\div$ denotes element-wise division and $\epsilon$ is set to $10^{-5}$. $O$ is the degraded image. 
We apply the loss function $L_{sup}(I,y)$ (i.e., Eq. \ref{eqlossres}) to supervise the training of the output variable $I$. where $y$ is the ground-truth clean image. The training data and settings of the U-Net are the same as those of RDM-IR. It should be noted that $I$ is only used in this training process.

For the auxiliary variables $P_0$ and $Q_0$, we directly set them to the same values as $T_0$ and $D_0$ as:
\begin{equation}
\begin{aligned}
P_0=T_0,Q_0=D_0.\label{eqinitpq}
\end{aligned}
\end{equation}

\subsection{Degradation Prior Transmitter}
\label{3.3}
\begin{figure}
	\begin{center}
		\includegraphics[width=\linewidth]{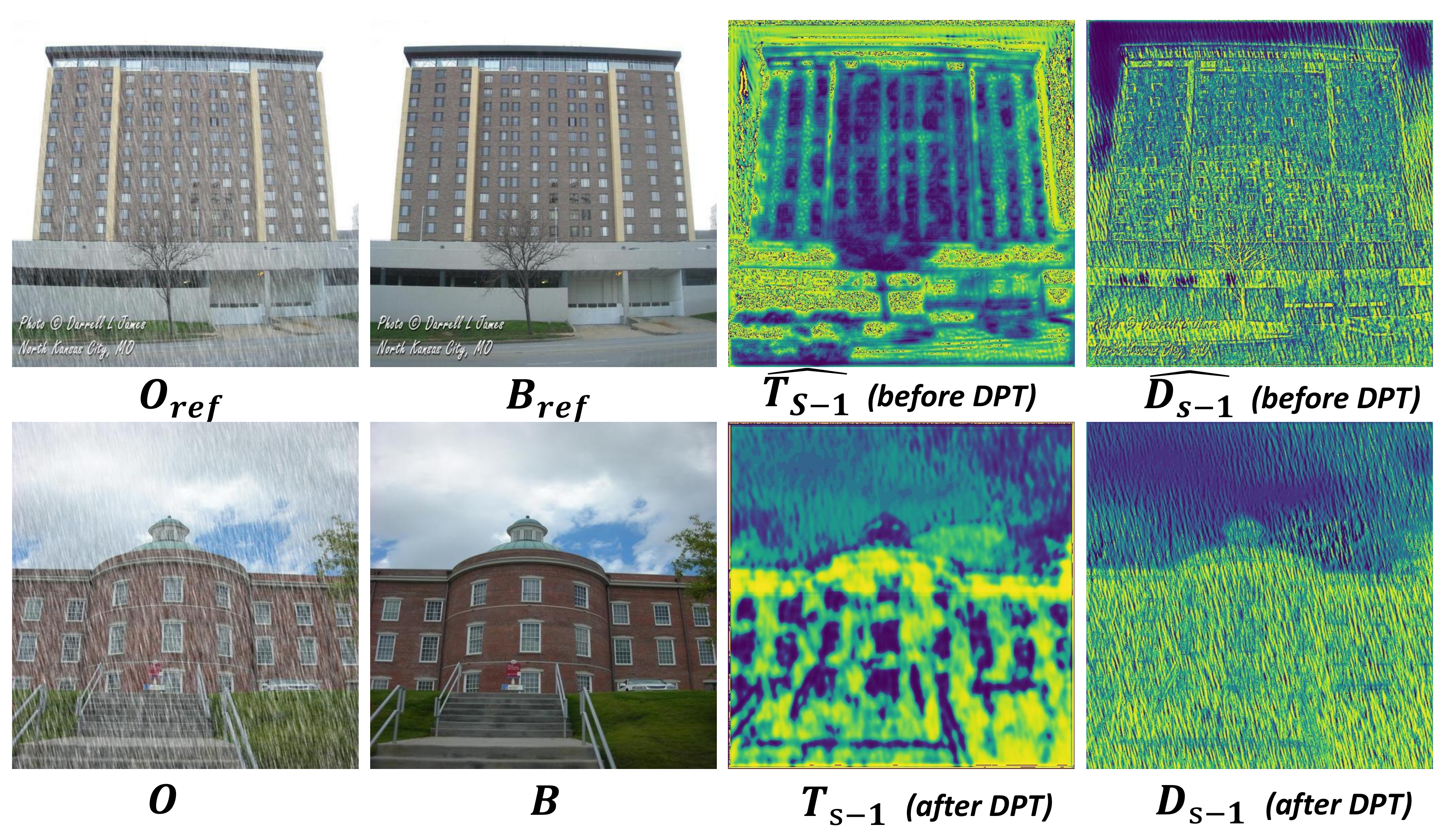}
	\end{center}
  \vspace{-10pt}
	\caption{Visualizations of the degradation matrices in final step $S$, which tend to be instance-dependent without DPT (top row). Please zoom in to see the details.}
	\label{imgfeat}
\end{figure}

The degradation matrices inevitably correlate strongly with the semantics of specific reference image instances.
Hence the generated degradation matrices ($\hat{T_k}$ and $\hat{D_k}$ in Eq. \ref{eqsoltdj1}) naturally contain \textbf{instance-dependent} information that corresponds with the reference image. These reference-oriented degradation matrices ($\hat{T_{s-1}}$ and $\hat{D_{s-1}}$ in the top row of Fig.~\ref{imgfeat}) potentially limit the restoration performance as the collected degradation statistics incompatible with the target image. 

Therefore, we further propose a DPT to extract more \textbf{instance-independent} information from the reference-oriented degradation matrices and couple it with the semantics of the target image ($T_{s-1}$ and $D_{s-1}$ in Fig. \ref{imgfeat}). 
As demonstrated in Fig. \ref{imgframework}(b), we use an encoder with CrossAttention~\cite{carion2020end} to extract instance-independent information and further produce the target-relevant degradation matrices that are semantically matched with the target degraded image. The updating process of the degradation matrices in DPT can be expressed as:
\begin{equation}
[T_k,D_k] = DPT([\hat{T_k},\hat{D_k}],B_k,[T_{k-1},D_{k-1}]),\label{eqdpt}
\end{equation}
where $[*]$ represents concatenation along the channel dimension. $[\hat{T_k},\hat{D_k}]$ is reference-oriented degradation matrices while $[T_k,D_k]$ denote the target-relevant matrices.

\subsection{Training and Loss Functions}
\label{3.4}
The training process of DRM-IR involves two stages. First, train a U-Net which is used to estimate the initial variables $T_0$ and $D_0$. Afterward, freeze the parameters of this U-Net and train the IR process in an end-to-end manner.
The supervising loss function $\mathcal{L}_{sup}$ is defined as the combination of SSIM loss \cite{wang1987ssim} and Charbonnier loss \cite{charbonnier1994two}:
\begin{equation}
\begin{aligned}
\mathcal{L}_{sup}(\hat{y},y) = (1-SSIM(\hat{y},y))+\sqrt{||\hat{y}-y||^2+\xi^2},
\end{aligned}
\end{equation}
where $\hat{y}$ is the predicted value and $y$ is the ground-truth value. The constant $\xi$ is empirically set to $10^{-3}$. 


\begin{figure*}[t]
	\begin{center}
		\includegraphics[width=\linewidth]{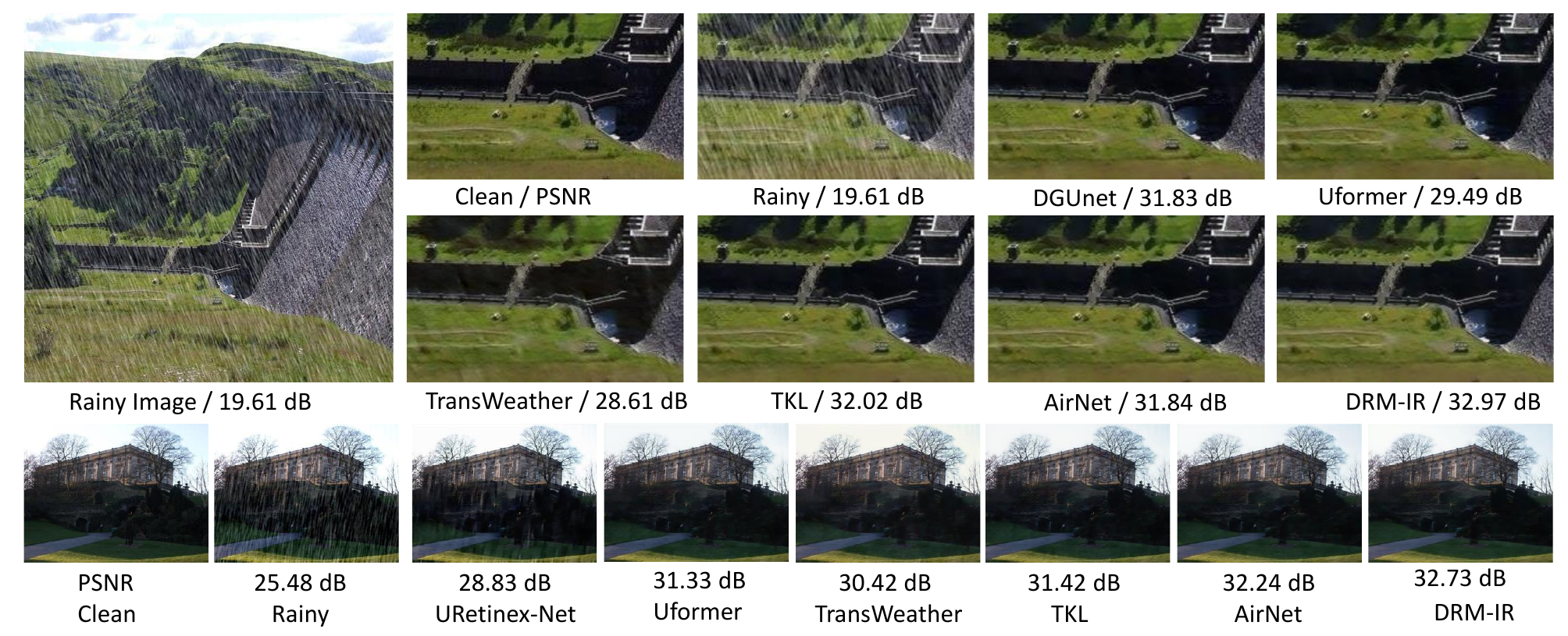}
	\end{center}
  \vspace{-10pt}
	\caption{Visual comparison of the deraining performance. Please zoom in for better visualization and comparison.}
	\label{imgderain}
\end{figure*}

\begin{figure*}[t]
	\begin{center}
		\includegraphics[width=\linewidth]{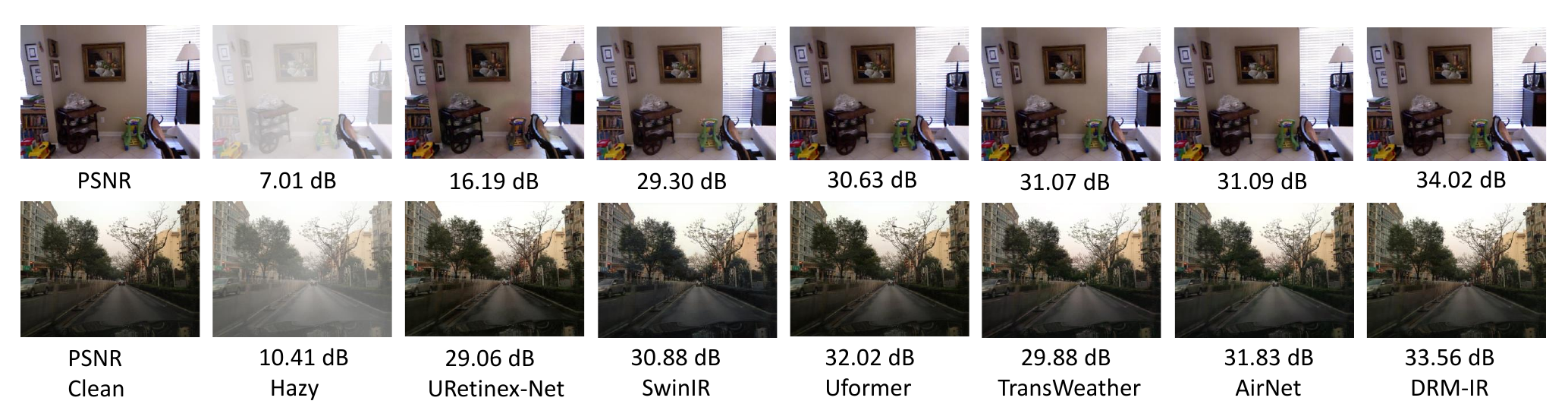}
	\end{center}
  \vspace{-10pt}
	\caption{Visual comparison of the dehazing performance. Please zoom in for better visualization and comparison.}
	\label{imgdehaze}
\end{figure*}

\begin{figure*}[t]
	\begin{center}
		\includegraphics[width=\linewidth]{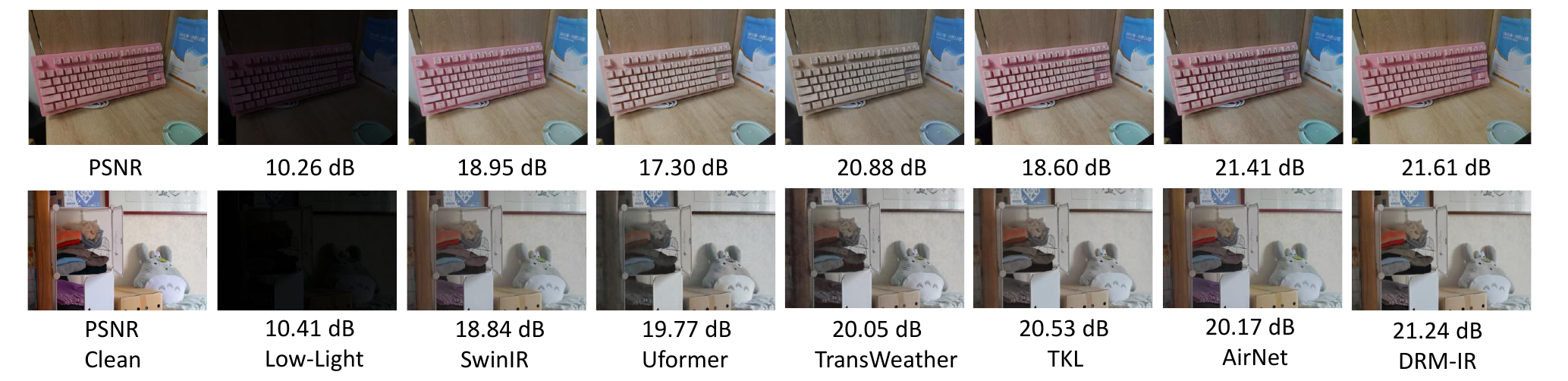}
	\end{center}
	\caption{Visual comparison of the low-light image enhancement performance. Please zoom in for better visualization and comparison.}
	\label{imgllie}
\end{figure*}

For the image restoring task, the outputs of each step are all supervised. Due to the step-wise iterative optimization, we apply exponentially growing weights to the loss function at each stage. Specifically, the supervised loss $\mathcal{L}_{res}$ for IR can be expressed as:
\begin{equation}
\begin{aligned}
\mathcal{L}_{res} = \sum_{k=1}^{S} w_k\mathcal{L}_{sup}(B_k,B),
\label{eqlossres}
\end{aligned}
\end{equation}
where $S$ is the total number of iteration steps and $w_k=\frac{2^k}{\sum_{i=1}^{S}2^i}$. Similarly, based on Eq. \ref{eqalldeg} and Eq. \ref{eqsoltdj1}, we present the loss function for the degradation modeling task as:
\begin{equation}
\begin{aligned}
\mathcal{L}_{deg} = \sum_{k=1}^{S-1} w_k\mathcal{L}_{sup}(\hat{T_k}B_{ref}+\hat{D_k},O_{ref}).
\label{eqlossdeg}
\end{aligned}
\end{equation}
Finally, the total loss function is the sum of $\mathcal{L}_{res}$ and $\mathcal{L}_{deg}$:
\begin{equation}
\begin{aligned}
\mathcal{L}_{total} = \mathcal{L}_{res}+\mathcal{L}_{deg}.
\label{eqlosstotal}
\end{aligned}
\end{equation}

\begin{figure*}[h]
	\begin{center}
		\includegraphics[width=\linewidth]{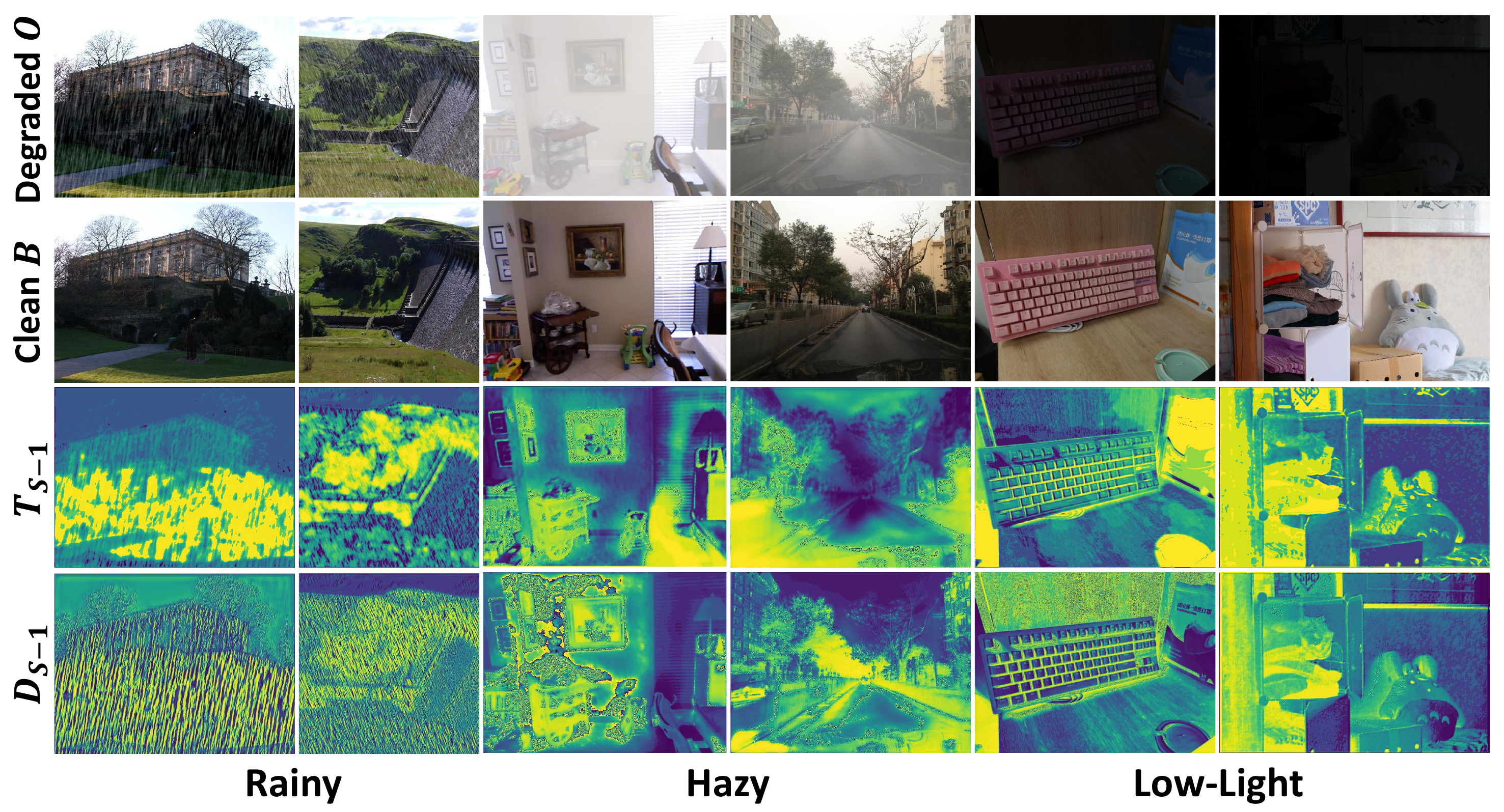}
	\end{center}
  \vspace{-10pt}
	\caption{Visualization of degradation matrices. A comparative analysis between Eq. \ref{eqalldeg} and Eq. \ref{eqrainy} and \ref{eqlowlight} shows that in rainy images, the variable T does not carry substantial significance, and so does the variable D in low-light images. This should be noticed in visual comparisons.}
	\label{imgvitds}
\end{figure*}

\section{Experiments and Analysis}
\label{4}

\noindent\textbf{Datasets.} The performance of All-In-One IR is evaluated on three common degradations (rain, haze, and low light). 
The constructed multi-degradation datasets Comb-1 and Comb-2 are detailed in Tab. \ref{tabdata}.

\begin{table}[h]
	\begin{center}
		\caption{Settings of the multi-degradation datasets.}\label{tabdata}
		\renewcommand\arraystretch{1.2}
			\tabcolsep=0.12cm
		\begin{tabular}{@{}|c|c|c|c|c|@{}}
			\hline
			& &Train&Test&\\
			\hline
			\multirow{3}*{Comb-1}&Rain&Rain12000&Test1200&\cite{zhang2018density}\\
			&Haze&ITS&SOTS-indoor&\cite{li2018benchmarking}\\
			&Low-Light &LOL-train&LOL-test&\cite{wei2018deep}\\
			\hline
			\multirow{3}*{Comb-2}&Rain&Rain800&Test100&\cite{zhang2019image}\\
			&Haze&OTS&SOTS-outdoor&\cite{li2018benchmarking}\\
			&Low-Light &LSRW-train&LSRW-test&\cite{hai2023r2rnet}\\
			\hline
		\end{tabular}
	\end{center}
\end{table}

\noindent\textbf{Training details.} 
The proposed DRM-IR is implemented by PyTorch 1.7.0 and all experiments in this work are conducted on an NVIDIA GeForce RTX 3090 GPU. Similar to previous work \cite{wu2022uretinex}, we set the initial values of the penalty parameters $\alpha$, $\beta$, and $\gamma$ to 0.5 and increased them by 0.05 at each step. The models are trained with Adam optimizers ($\beta_1=0.9$, $\beta_2=0.999$, $weight\_decay =10^{-4}$) for 400K iterations with $batch\_size=4$. The initial value of the learning rate is set to $2\times 10^{-4}$ and tenfold reduction every 150K iterations. It should be noticed that there is a considerable difference in the number of images in different datasets. In order to avoid the long-tailed distribution of training data interfering with the model training, we randomly sample rainy, hazy, and low-light training images with equal probability in the training process. Besides, The reference image pairs are randomly sampled from the training set and have the same kinds of degradations as the training images. The scales of both the training and reference images are resized to $256\times 256$ during the training process.

\noindent\textbf{Evaluating details.} 
Following the previous works \cite{zamir2022restormer,wu2021contrastive,ma2022toward}, we evaluate the IR performance of different methods in terms of the Peak Signal-to-Noise Ratio (PSNR) and the structural similarity (SSIM) \cite{wang1987ssim}. 
Results in all following tables are reported as the average of the scores with ten different groups of referenced images unless indicated specifically. 
Note that for a fair comparison with existing methods that do not require a type prior, in this work, our degradation types are obtained from a pre-trained ResNet-18~\cite{he2016deep} classifier instead of being manually specified.

\begin{table*}
	\begin{center}
		\caption{Quantitative comparison on All-In-One IR. The best and second-best performance are \textbf{highlighted} and \underline{underlined}.}\label{taballinone}
		\renewcommand\arraystretch{1.15}
		\tabcolsep=0.25cm
		\begin{tabular}{@{}c|cc|cccccc|cc@{}}
			\hline
			\hline
			\multirow{2}*{Dataset}&\multirow{2}*{Method}&\multirow{2}*{Type}&\multicolumn{2}{c}{Deraing}&\multicolumn{2}{c}{Dehazing}&\multicolumn{2}{c}{LLIE}&\multicolumn{2}{c}{Average}\\
			& & & PSNR$\uparrow$&SSIM$\uparrow$&PSNR$\uparrow$&SSIM$\uparrow$&PSNR$\uparrow$&SSIM$\uparrow$&PSNR$\uparrow$&SSIM$\uparrow$\\
			\hline
			\hline
			\multirow{9}*{Comb-1}&Zhang \emph{et al.}&\multirow{3}*{Unfolding-based}&26.354&0.827&28.832&0.940&20.882&0.823&25.356&0.863\\
			&URetinex-Net& &26.777&0.831&30.671&0.973&21.148&0.818&26.199&0.874\\
			
			&DGUNet& &31.449&0.912&34.780&0.979&21.201&0.819&29.160&0.903\\
			
			\cline{4-11}
			&SwinIR&\multirow{2}*{Task-specific}&32.610&0.917&33.947&0.976&21.139&0.819&29.232&0.903\\
			&Uformer& &32.566&0.913&35.452&0.981&21.201&0.825&29.740&0.906\\
			\cline{4-11}
			&TransWeather&\multirow{3}*{All-In-One}&31.049&0.896&35.346&0.965&21.143&0.819&29.179&0.893\\
			&TKL& &\underline{32.803}&\underline{0.921}&35.725&0.986&21.204&0.827&29.911&0.911\\
			&AirNet& &32.397&0.920&\underline{36.990}&\underline{0.987}&\underline{21.206}&\underline{0.828}&\underline{30.198}&\underline{0.912}\\
			\cline{4-11}
			&DRM-IR&--- &\textbf{32.965}&\textbf{0.925}&\textbf{37.185}&\textbf{0.994}&\textbf{21.309}&\textbf{0.836}&\textbf{30.486}&\textbf{0.918}\\
			\hline
			\hline
			\multirow{9}*{Comb-2}&Zhang \emph{et al.}&\multirow{3}*{Unfolding-based}&25.016&0.828&27.542&0.938&15.330&0.458&22.629&0.741\\
			&URetinex-Net& &25.464&0.841&27.839&0.955&16.601&0.501&23.301&0.766\\
			
			&DGUNet& &30.576&0.902&28.209&0.957&17.184&0.522&25.323&0.794\\
			
			\cline{4-11}
			&SwinIR&\multirow{2}*{Task-specific}&30.996&0.909&28.103&0.947&16.329&0.477&25.143&0.778\\
			&Uformer& &31.602&0.912&28.044&0.944&17.034&0.511&25.561&0.789\\
			\cline{4-11}
			&TransWeather&\multirow{3}*{All-In-One}&31.044&0.911&27.826&0.949&17.531&0.533&25.467&0.798\\
			&TKL& &31.763&\underline{0.917}&28.165&0.953&\underline{17.546}&\underline{0.540}&25.824&0.803\\
			&AirNet& &\underline{31.825}&0.915&\underline{28.297}&\underline{0.962}&17.535&0.534&\underline{25.886}&\underline{0.804}\\
			\cline{4-11}
			&DRM-IR&--- &\textbf{31.902}&\textbf{0.919}&\textbf{29.436}&\textbf{0.968}&\textbf{17.582}&\textbf{0.543}&\textbf{26.307}&\textbf{0.810}\\
			\hline
			\hline
		\end{tabular}
	\end{center}
\end{table*}

\begin{table*}
	\begin{center}
		\caption{Quantitative comparison on Task-specific IR. The best and second-best performance are \textbf{highlighted} and \underline{underlined}.}\label{tabtaskspec}
		\renewcommand\arraystretch{1.15}
		\tabcolsep=0.3cm
		\begin{tabular}{@{}c|cccccc|c@{}}
			\hline
			\hline
\multirow{2}*{Method}&\multicolumn{2}{c}{Deraing}&\multicolumn{2}{c}{Dehazing}&\multicolumn{2}{c}{LLIE}&\multirow{2}*{Average}\\
&Test100&Test1200&SOTS-indoor&SOTS-outdoor&LOL-test&LSRW-test&\\
			\hline
			\hline
SwinIR&31.11 / 0.913&32.73 / 0.919&35.02 / 0.955&28.71 / 0.931&20.78 / 0.825&16.51 / 0.507&27.48 / 0.842\\
MPRNet&30.27 /  0.897&32.91 / 0.916&35.47 / 0.963&28.76 / 0.948&20.99 / 0.842&17.08 / 0.519&27.58 / 0.848\\
DGUNet&30.86 /  0.907&\textbf{33.08} /  0.916&37.23 / 0.985&\underline{29.11} / \underline{0.964}&21.33 / \underline{0.847}&17.39 / \underline{0.553}&28.17 / 0.862\\
Uformer&\textbf{31.95} / \textbf{0.922}&32.69 / \underline{0.920}&\underline{37.43} / \underline{0.993}&29.01 / 0.959&\underline{21.59} / 0.846&\underline{17.47} / 0.536&\underline{28.36} / \underline{0.863}\\
DRM-IR&\underline{31.94} / \underline{0.920}&\underline{33.05} / \textbf{0.923}&\textbf{38.41} / \textbf{0.996}&\textbf{30.07} / \textbf{0.974}&\textbf{21.67} / \textbf{0.852}&\textbf{17.83} / \textbf{0.573}&\textbf{28.83} / \textbf{0.873}\\
			\hline
			\hline
		\end{tabular}
	\end{center}
\end{table*}

\subsection{Comparison with state-of-the-art methods on All-In-One IR}
\label{4.1}
The performance comparison with state-of-the-art methods is conducted on Comb-1 and Comb-2 datasets. Among them, Zhang \emph{et al.}'s work \cite{zhang2017learning}, DGUNet\cite{mou2022deep} and URetinex-Net \cite{wu2022uretinex} are Deep Unfolding-based methods; SwinIR \cite{liang2021swinir} and Uformer \cite{wang2022uformer} are task-specific IR methods; TKL \cite{chen2022learning}, TransWeather \cite{valanarasu2022transweather}, and AirNet \cite{li2022all} are recently proposed All-In-One IR methods. All the models are trained and evaluated in the same experimental setup for a fair comparison.
Fig. \ref{imgderain}, \ref{imgdehaze} and \ref{imgllie} demonstrates the visual comparisons of DRM-IR with several recent methods in deraining, dehazing, and low-light-image-enhancement.
The quantitative scores of each model are reported in Tab. \ref{taballinone}. 
It can be observed that the proposed DRM-IR consistently outperforms both task-specific and All-In-One competitors across various degradation scenarios.

\subsection{Comparison with state-of-the-art methods on Task-specific IR}
We also conducted quantitative comparisons with state-of-the-art methods in task-specific IR, including SwinIR\cite{liang2021swinir}, MPRNet\cite{mehri2021mprnet}, DGUNet\cite{mou2022deep}, and Uformer\cite{wang2022uformer}. The experiments in this study were conducted with the same experimental settings as existing methods to facilitate a fair and quantitative comparison. Tab. \ref{tabtaskspec} presents the quantitative evaluation results for deraing, dehazing, and LLIE of each method. The proposed DRM-IR outperforms existing methods on the majority of datasets. This demonstrates that DRM-IR, in addition to its dynamic advantages in the All-In-One scenario, further improves image restoration accuracy through the cascaded paradigm of modeling and restoration.

\subsection{Ablation Study}
\label{4.3}
In this section, we discuss the impacts of several critical designs in DRM-IR in terms of the All-In-One IR performance. All the ablation studies are conducted on the Comb-1 dataset with performance evaluated using PSNR.

\begin{table}
	\begin{center}
		\caption{Ablation on the reference-based degradation modeling. The model with extra references achieves preferable results.}
		\label{tabref}
		\renewcommand\arraystretch{1.15}
		\tabcolsep=0.3cm
		\begin{tabular}{@{}ccccc@{}}
			\hline
			\hline
			Method&Deraining&Dehazing&LLIE&Average\\
			\hline
			$w/o$ ref&32.217&36.535&21.027&29.926\\
			$w/$ ref&\textbf{32.965}&\textbf{37.185}&\textbf{21.309}&\textbf{30.486}\\
			\hline\hline
		\end{tabular}
	\end{center}
\end{table}

\begin{figure}
	\begin{center}
		\includegraphics[width=\linewidth]{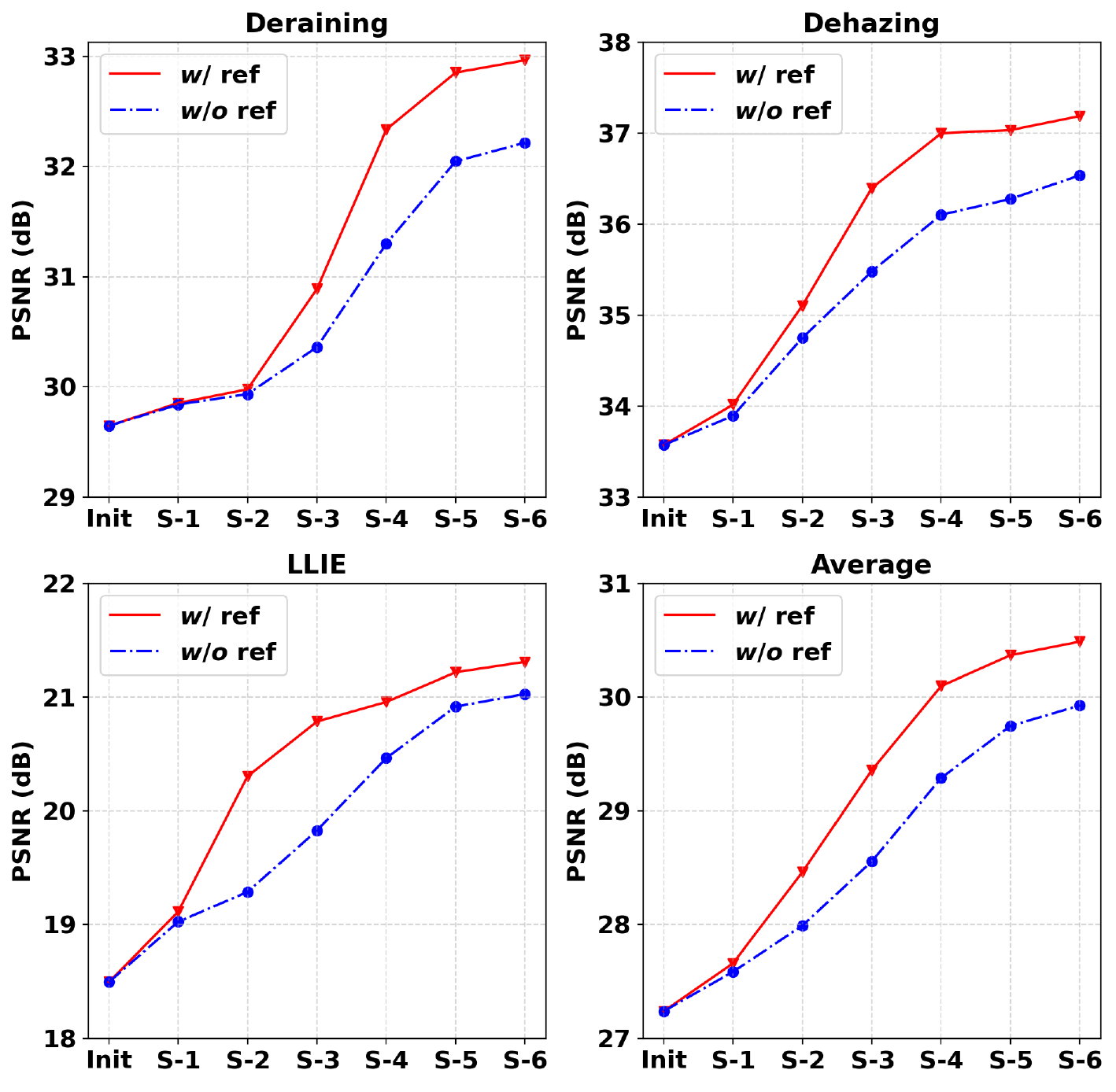}
	\end{center}
  \vspace{-15pt}
	\caption{PSNR scores of the outputs $B_k$ for each step with and without reference-based degradation modeling. $Init$ represents the cursory clean image $I$ estimated by $I=(O-D_0)/(T_0+\epsilon)$, where the hyper-parameter $\epsilon$ is set as $10^{-5}$. S-$i$ represents the $i$-th step}
	\label{imgline}
\end{figure}

\noindent\textbf{Reference-based task-adaptive modeling.} This work introduces additional degradation information from reference image pairs and updates the degradation matrices for more precise modeling. 
For instance, it effectively captures rain streaks in rainy images, depth information in hazy images, and darkness in low-light images (see Fig. \ref{imgvitds}).
Tab. \ref{tabref} shows the performance comparison of the proposed method with and without the presence of the reference-based degradation modeling. 
As for the variant without the reference-based degradation modeling, the degradation matrices are fixed as the initial value. 
It can be observed from Tab. \ref{tabref} that the IR performance achieves a significant improvement with the proposed reference-based method. 
Moreover, the PSNR performance of each iteration step is illustrated in Fig. \ref{imgline}. It can be observed that our reference-based modeling mechanism not only improves the performance but also facilitates iteration efficiency.

\begin{figure*}
	\begin{center}
		\includegraphics[width=\linewidth]{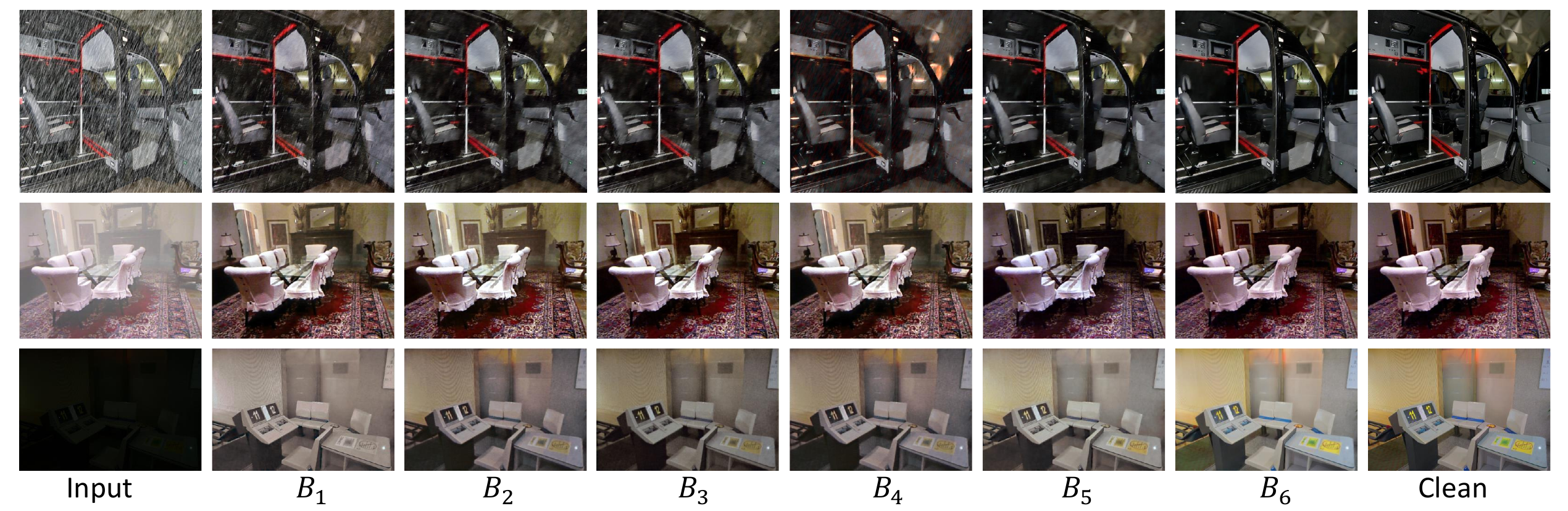}
	\end{center}
  \vspace{-13pt}
	\caption{Visualizations of $B_k$ for each step during the IR process.}
	\label{imgresprocess_l}
\end{figure*}

\noindent\textbf{Iteration number of the unfolding optimization.}
From the results reported in Tab. \ref{tabit}, the proposed method achieves state-of-the-art performance after $6$ iterations. Therefore, considering the computational cost and restoring performance comprehensively, the number of iterations is set to $6$ in this paper. Fig. \ref{imgresprocess_l} further illustrates the qualitative restoration processes of the degraded images. As the optimization proceeds, the degradations are progressively eliminated and satisfactory results are finally obtained after $6$ iterations.

\begin{table}
	\begin{center}
		\caption{Ablation study on the number of unfolding iterations. As the total number of iterations increases, both the computational cost and the performance progressively rise. DRM-IR achieves state-of-the-art performance after 6 iterations.}
		\tabcolsep=0.1cm
		\label{tabit}
		\renewcommand\arraystretch{1.1}
		\begin{tabular}{@{}ccccccc@{}}
			\hline
			\hline
		Number&Deraining&Dehazing&LLIE&Average&GMacs&Time(s)\\
			\hline
			1&29.123&31.973&19.150&26.749&14.38&0.063\\
			2&29.617&33.129&19.185&27.310&31.35&0.083\\
			3&30.576&34.991&20.266&28.611&48.32&0.095\\
			4&32.183&36.228&21.107&29.839&65.29&0.109\\
			5&32.799&36.981&21.232&30.337&82.26&0.125\\
			6&\textbf{32.965}&\textbf{37.185}&\textbf{21.309}&\textbf{30.486}&99.23&0.142\\
			\hline\hline
		\end{tabular}
	\end{center}
\end{table}

\begin{table}
	\begin{center}
		\caption{Ablation results on DPT. $Cat$ indicates replacing the CorssAttention block in Fig. \ref{imgframework} with the Concatenation operation.}
		\tabcolsep=0.22cm
		\label{tabdpt}
		\renewcommand\arraystretch{1.1}
		\begin{tabular}{@{}ccccc@{}}
			\hline
			\hline
			Method&Deraining&Dehazing&LLIE&Average\\
			\hline
			Cat&28.398&31.463&18.140&26.000\\
			Full DPT&\textbf{32.965}&\textbf{37.185}&\textbf{21.309}&\textbf{30.486}\\
			\hline\hline
		\end{tabular}
	\end{center}
\end{table}
\noindent\textbf{Degradation Prior Transmitter.} 
Since the degradation information is deeply entangled with the image semantics (\emph{e.g.}, haze concentration is related to the scene depth~\cite{laina2016deeper,ranftl2020towards,yang2022self}), degradation matrices obtained from the reference image pair inevitably contains instance-specific statistics (top row of Fig. \ref{imgfeat}) and cannot be directly used for the target image restoration. Therefore, DPT is devised to transfer more universal degradation information to the target degraded image. 
Experimental results in Tab. \ref{tabdpt} demonstrate that DPT is essential for better utilizing the additional degradation priors, which cooperates well with the proposed reference-based task-adaptive modeling paradigm.


\begin{table}
	\begin{center}
		\caption{Performance comparisons of transmitting degradation information with feature maps from different depths. $2\times$ indicates the corresponding layer after 2 times downsampling, and so on.}
		\label{tabtrans}
		\tabcolsep=0.32cm
		\renewcommand\arraystretch{1.2}
		\begin{tabular}{@{}ccccc@{}}
			\hline
			\hline
			Scale&Deraining&Dehazing&LLIE&Average\\
			\hline
			$2\times$&29.493&33.086&19.277&27.285\\
			$4\times$&29.950&34.491&20.259&28.233\\
			$8\times$&31.080&35.371&20.306&28.919\\
			$16\times$&\textbf{32.965}&\textbf{37.185}&\textbf{21.309}&\textbf{30.486}\\
			\hline\hline
		\end{tabular}
	\end{center}
\end{table}

\begin{figure}
	\begin{center}
		\includegraphics[width=\linewidth]{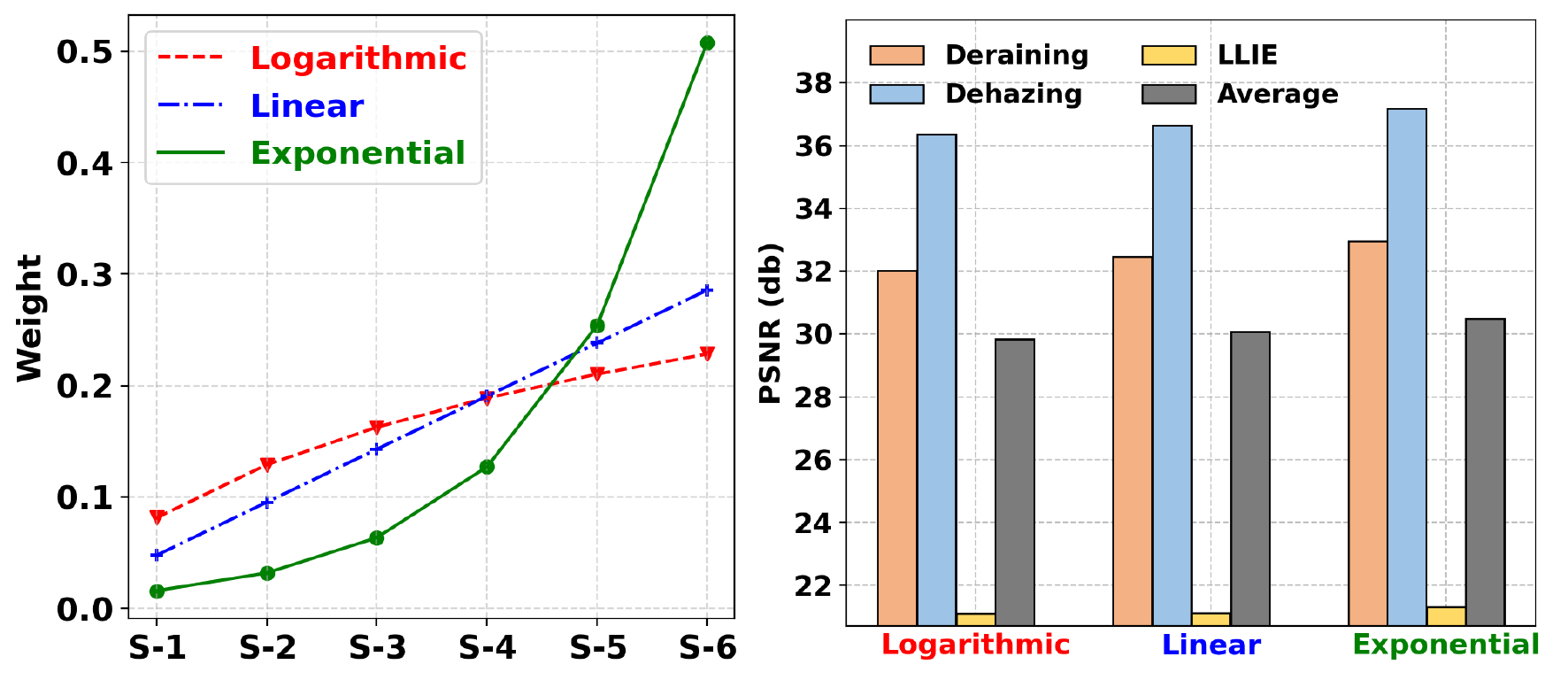}
	\end{center}
  \vspace{-13pt}
	\caption{Comparisons results on different weight growth rates during the optimization, as well as the image restoration performance at each rate. Please zoom in to see the details. S-$i$ represents the $i$-th step}
	\label{imgweight}
\end{figure}

\noindent\textbf{Loss function weights for each step.} The output image $B_k$ should be progressively restored during the unfolding iteration. Hence we increase the weights of the loss functions along with the iterations. As shown in Tab. \ref{tabweights} and Fig. \ref{imgweight}, we explored three growth rates including logarithmic growth, linear growth, and exponential growth. Among them, exponential increment presented the optimum performance.
\begin{table}
	\begin{center}
		\caption{The weights of the loss functions imposed on each step of the unfolding iterative process. The exponentially increasing weights achieved the best results. In practice, the total number of unfolding iterations $S$ is set to 6.}
		\label{tabweights}
		\renewcommand\arraystretch{1.8}
		\tabcolsep=0.1cm
		\begin{tabular}{@{}ccccc@{}}
			\hline
			\hline
                Weight&Deraining&Dehazing&LLIE&Average\\
			\hline
			$w_k$=$\frac{\log_2{(k+1)}}{\sum_{i=1}^{S}\log_2(i+1)}$&32.027&36.364&21.086&29.826\\
			$w_k$=$\frac{k}{\sum_{i=1}^{S}i}$&32.466&36.649&21.113&30.076\\
			$w_k$=$\frac{2^k}{\sum_{i=1}^{S}2^i}$&\textbf{32.965}&\textbf{37.185}&\textbf{21.309}&\textbf{30.486}\\
			\hline\hline
		\end{tabular}
	\end{center}
\end{table}

\begin{figure}[h]
	\begin{center}
		\includegraphics[width=\linewidth]{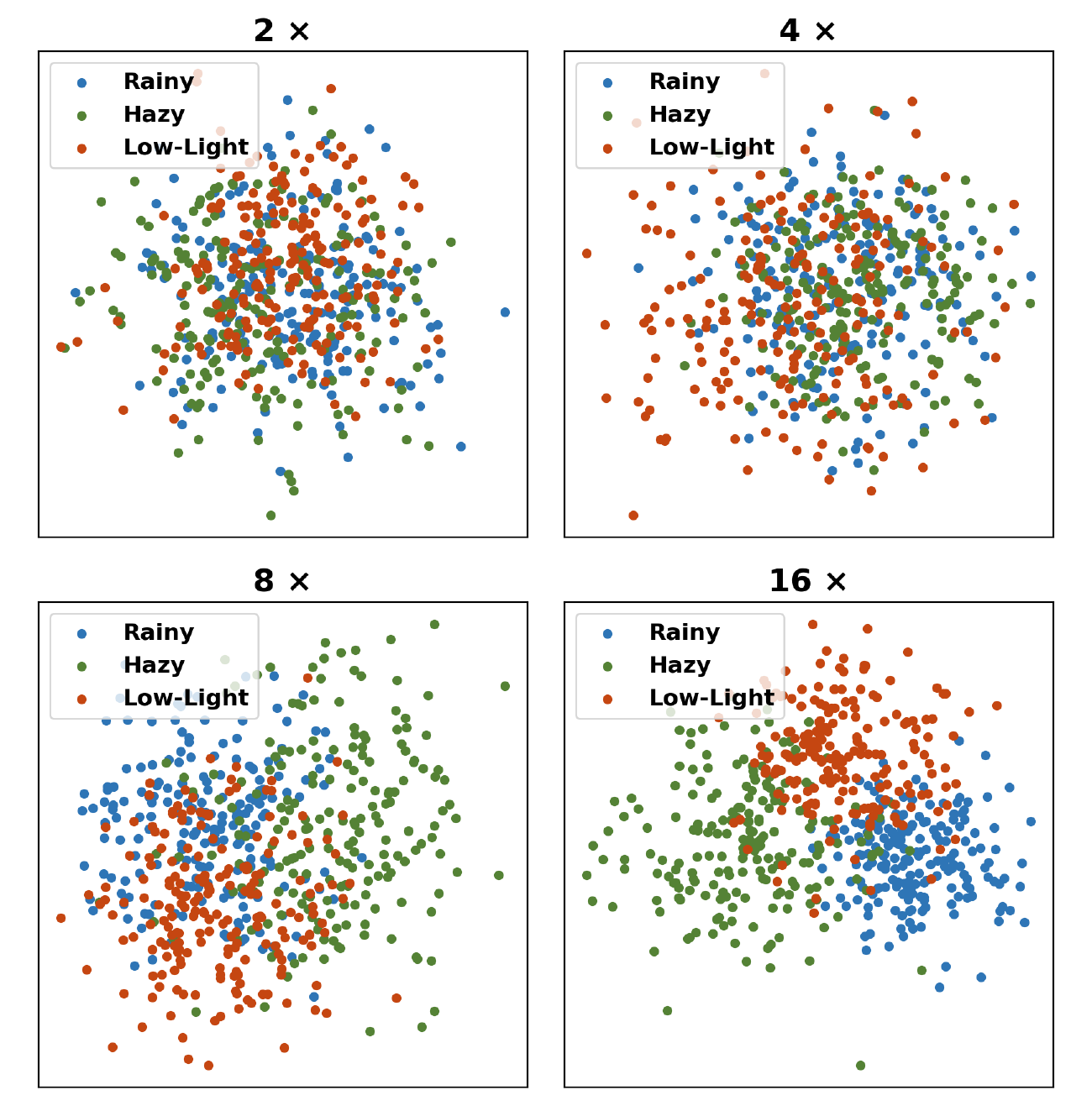}
	\end{center}
  \vspace{-13pt}
	\caption{t-SNE visualizations of feature maps in different layers. $2\times$ in the figure indicates 2 times downsampling, and so on.}
	\label{imgtsne}
\end{figure}

\noindent\textbf{Sampling strategy in DPT }.
As described in Sec. \ref{3.3} and Fig. \ref{imgframework}(b) in the paper, the transmitted degradation information in DPT come from a certain layer of the encoder.
To obtain more instance-independent features in the reference images, we quantitatively prove that deeper features contain ``purer'' general information through the ablations on different sampling depths in Tab. \ref{tabtrans}.
In addition, We randomly selected 200 images in each of the three categories of rainy, hazy, and low-light images for qualitative comparison.
As shown in Fig. \ref{imgtsne}, we further adopt t-SNE~\cite{t-SNE} on the feature maps in different layers of the DPT encoder. It can be clearly observed that the deepest feature map after $16 \times$ downsampling contains fewer instance-specific features and contains purer instance-independent universal degenerate features. 

\noindent\textbf{Degradation modeling form}. 
In the previous literature \cite{zhang2017learning, jin2022shadowdiffusion}, unfolding-based methods are usually based on modeling with the form of $O=HB$, where $H$ is the degradation matrix. 
In contrast, since our DRM-IR is devoted to the All-In-One image restoration problem, a generalized degradation model (i.e., $O=TB+D$) is proposed. 
To verify the effectiveness of the proposed modeling form, we also conducted relevant experiments. As shown in Tab. \ref{tabmod}, it is evident that the proposed generalized degradation modeling form is superior for the All-In-One image restoration task.

\begin{table}
	\begin{center}
\caption{Ablation study for degradation modeling form. The proposed generalized degradation modeling formula is more appropriate for All-In-One image restoration.}
\tabcolsep=0.02cm
\label{tabmod}
\renewcommand\arraystretch{1.3}
	\tabcolsep=0.15cm
\begin{tabular}{@{}ccccc@{}}
	\hline
	\hline
	Method&Deraining&Dehazing&LLIE&Average\\
	\hline
	$O=HB$&30.483&34.910&21.022&28.805\\
	$O=TB+D$&\textbf{32.965}&\textbf{37.185}&\textbf{21.309}&\textbf{30.486}\\
	\hline\hline
\end{tabular}
	\end{center}
\end{table}

\begin{table}
	\begin{center}
		\caption{All-In-One IR performance comparison between parallel and serial architectures.}
		\label{tabpors}
		\renewcommand\arraystretch{1.15}
		\tabcolsep=0.3cm
		\begin{tabular}{@{}ccccc@{}}
			\hline
			\hline
			Architecture&Deraining&Dehazing&LLIE&Average\\
			\hline
			Serial&32.194&36.807&21.009&30.003\\
			Parallel&\textbf{32.965}&\textbf{37.185}&\textbf{21.309}&\textbf{30.486}\\
			\hline\hline
		\end{tabular}
	\end{center}
\end{table}

\noindent\textbf{Parallel or serial?}
For the tasks-adaptive degradation modeling and model-based image restoring, we explored both parallel and serial architectures. The parallel architecture, as described in Sec. \ref{3}, was ultimately chosen as the preferred framework. In the serial architecture, the process involved first optimizing Eq. \ref{eqenergyfdeg} to obtain degradation matrices, and then inputting the obtained degradation matrices into each stage of the optimization process in Eq. \ref{eqenergyfres}. Each of the two optimization processes underwent $6$ iterations. Tab. \ref{tabpors} presents the final results for the two architectures. The parallel architecture outperforms the serial architecture. This is because in each stage of the parallel architecture, the inaccuracy of the optimized variables (degradation matrices or clean background) not only leads to a larger value of the corresponding task's energy function (Eq. \ref{eqenergyfdeg} or Eq. \ref{eqenergyfres}) but also raises the value of the other task's energy function (Eq. \ref{eqenergyfres} or Eq. \ref{eqenergyfdeg}). As a result, the optimization of both tasks in the parallel architecture mutually reinforces each other.

\section{Conclusion}
\label{6}
This paper studies the unfolding-based All-In-One image restoration. 
Combing the flexibility of model-based methods with the portability of learning-based methods, we innovatively propose a flexible Dynamic Reference Modeling paradigm, namely DRM-IR, with two cascaded MAP inferences respectively focusing on task-adaptive degradation modeling and model-based image restoring. 
In particular, we first introduce a reference-based mechanism for precise task-adaptive degradation modeling. Coordinated with the subsequent model-based restoring process, the proposed DRM-IR is able to uniformly remove various degradations with an efficient framework while being interpretable. Besides, a degradation prior transmitter (DPT) is further introduced to entangle the cascaded modeling and restoring processes. 
Comprehensive qualitative and quantitative experiments demonstrate that our DRM-IR achieves state-of-the-art performance for All-In-One image restoration.

\bibliographystyle{IEEEtran}
\bibliography{egbib}


 

\begin{IEEEbiography}[{\includegraphics[width=1in,height=1.25in,clip,keepaspectratio]{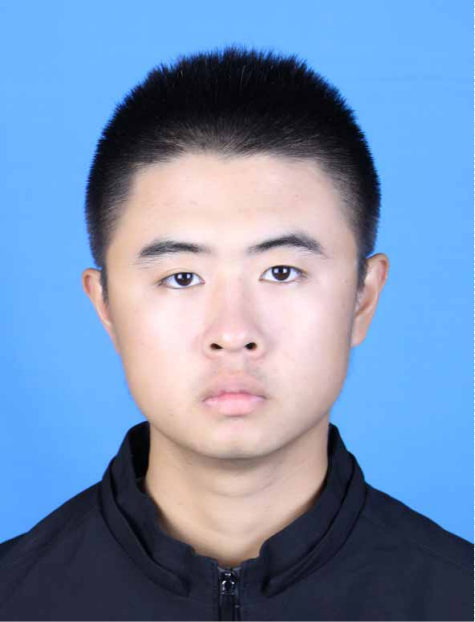}}]{Yuanshuo Cheng} received the B.E. degree in College of Computer Science and Technology, China University of Petroleum (East China), in 2023. He is currently working toward the M.S. degree in College of Computer Science and Technology, China University of Petroleum (East China) under the supervision of Prof. Shao. His current research interests include image restoration, computer vision, and deep learning.
\end{IEEEbiography}
\vspace{-15pt}
\begin{IEEEbiography}[{\includegraphics[width=1in,height=1.25in,clip,keepaspectratio]{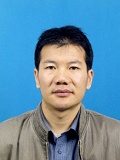}}]{Mingwen Shao} received his M.S. degree in mathematics from the Guangxi University, Guangxi, China, in 2002, and the Ph.D. degree in applied mathematics from Xi'an Jiaotong University, Xi'an, China, in 2005. He received the postdoctoral degree in control science and engineering from Tsinghua University in February 2008. Now he is a professor and doctoral supervisor at China University of Petroleum (East China). His research interests include machine learning, computer vision, and data mining.
\end{IEEEbiography}
\vspace{-15pt}
\begin{IEEEbiography}[{\includegraphics[width=1in,height=1.25in,clip,keepaspectratio]{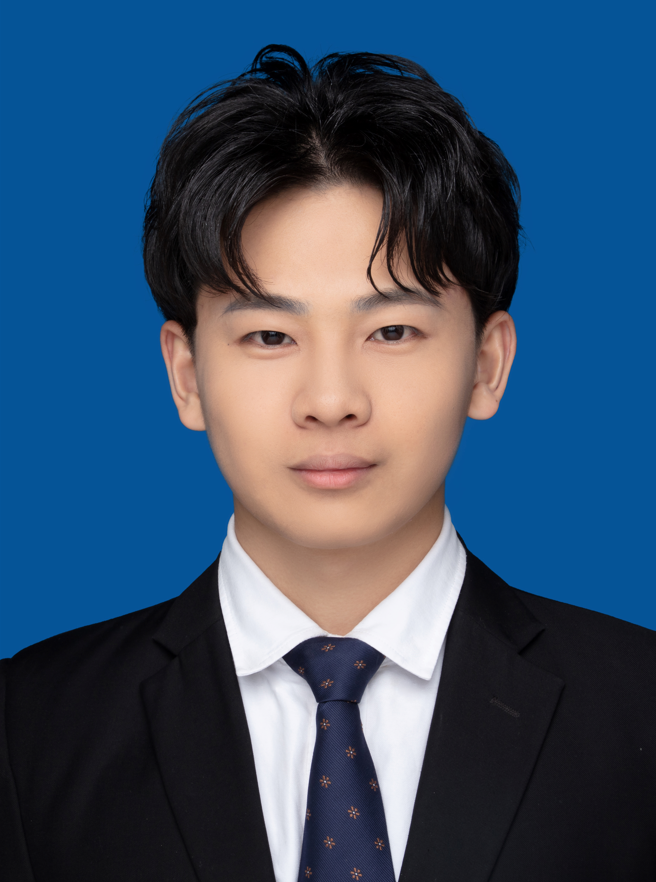}}]{Yecong Wan} received the B.E. degree in College of Computer Science and Technology, China University of Petroleum (East China), in 2023. He is currently working toward the M.S. degree in College of Computer Science and Technology, China University of Petroleum (East China) under the supervision of Prof. Shao. His current research interests include image restoration and computer vision. 
\end{IEEEbiography}
\vspace{-15pt}
\begin{IEEEbiography}[{\includegraphics[width=1in,height=1.25in,clip,keepaspectratio]{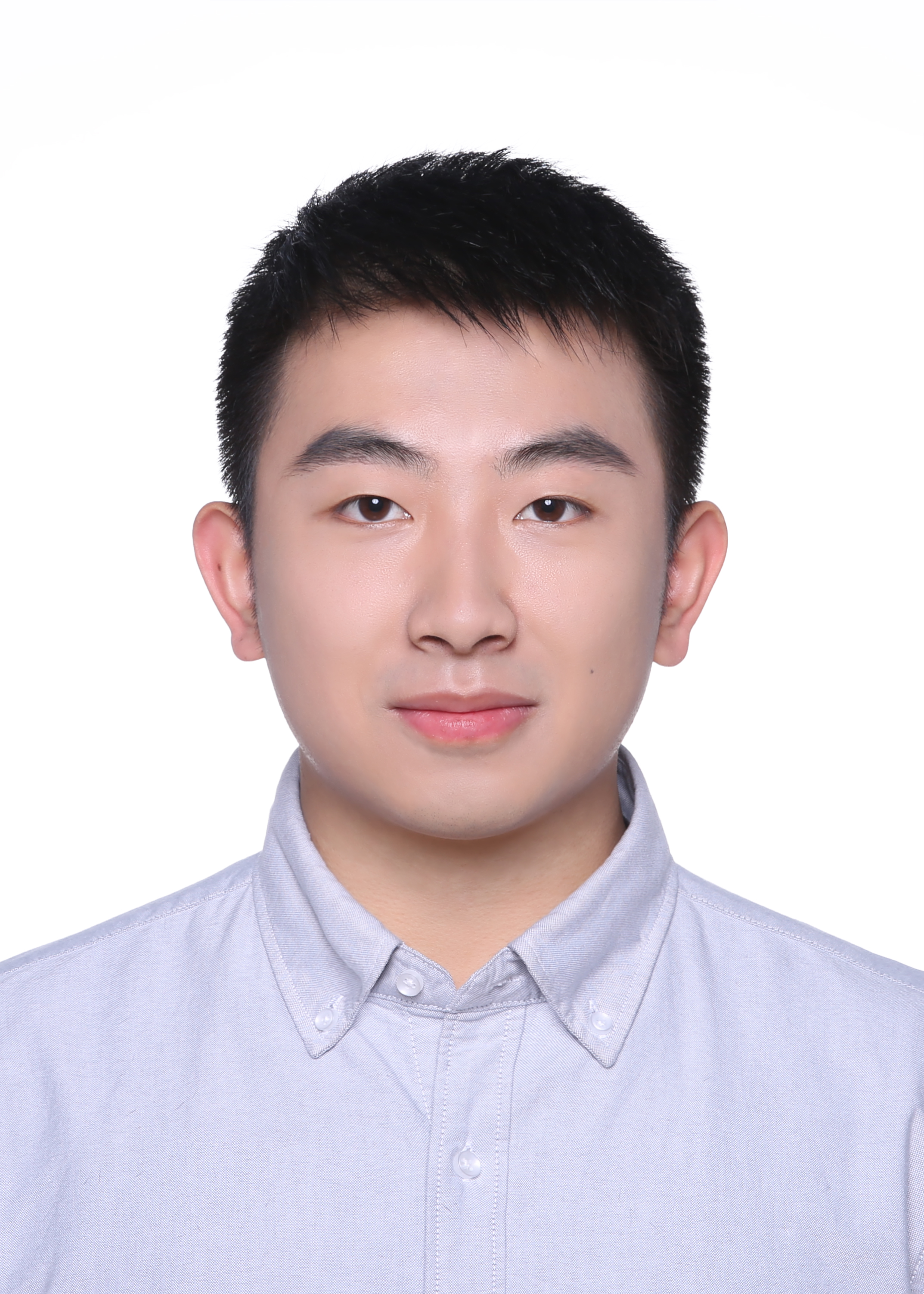}}]{Chao Wang} received the M.S. degree from China University of Petroleum, City Qingdao, China, in 2022. He is currently pursuing the Ph.D. degree with the University of Technology Sydney, Australia. His research interests
include image synthesis and restoration.
\end{IEEEbiography}
\vspace{-15pt}


\vfill

\end{document}